\documentclass[10pt,twocolumn,letterpaper]{article}

\usepackage[pagenumbers]{cvpr} 

\usepackage{graphicx}
\usepackage{amsmath}
\usepackage{amssymb}
\usepackage{booktabs}
\usepackage{multirow}

\usepackage[pagebackref,breaklinks,colorlinks]{hyperref}

\usepackage[capitalize]{cleveref}
\crefname{section}{Sec.}{Secs.}
\Crefname{section}{Section}{Sections}
\Crefname{table}{Table}{Tables}
\crefname{table}{Tab.}{Tabs.}

\def\confName{CVPR}
\def\confYear{2023}

\hypersetup{citecolor=.}

\usepackage{commenting}
\usepackage{microtype}          
\usepackage{booktabs}
\usepackage{tablefootnote}
\usepackage{xspace}
\newcommand{\eug}{\mbox{e-UG}\xspace}
\newcommand{\gpt}{\mbox{GPT-2}\xspace}
\newcommand{\nlxgpt}{\mbox{NLX-GPT}\xspace}
\newcommand{\ofaxmt}{\mbox{OFA-X\textsubscript{\textsf{MT}}}\xspace}

\begin{document}

\title{Harnessing the Power of Multi-task Pre-training for Ground-truth-level \\
Natural Language Explanations}

\author{Björn Plüster$^{*1}$, Jakob Ambsdorf$^{*\dagger2}$, Lukas Braach$^{1}$, Jae Hee Lee$^{1}$, Stefan Wermter$^{1}$\\
  $^{1}$University of Hamburg, Germany\\
  $^{2}$University of Copenhagen, Denmark\\ 
   {\tt\small \{bjoern.pluester, lukas.braach\}@studium.uni-hamburg.de}\\
   {\tt\small jaam@di.ku.dk} \\
    {\tt\small \{jae.hee.lee, stefan.wermter\}@uni-hamburg.de }
  }
\maketitle

\iftoggle{cvprfinal}{%
   \def\thefootnote{*}\footnotetext{These authors contributed equally to this work}
   \def\thefootnote{$\dagger$}\footnotetext{Majority of work done while at University of Hamburg}
   \def\thefootnote{}\footnotetext{Code and demo available at \href{https://github.com/ofa-x/OFA-X}{https://github.com/ofa-x/OFA-X}}
}

\begin{abstract}
Natural language explanations promise to offer intuitively understandable explanations of a neural network's decision process in complex vision-language tasks, as pursued in recent VL-NLE research. While current models offer impressive performance on task accuracy and explanation plausibility, they suffer from a range of issues: Some models feature a modular design where the explanation generation module is poorly integrated with a separate module for task-answer prediction, employ backbone models trained on limited sets of tasks, or incorporate ad hoc solutions to increase performance on single datasets. We propose to avoid these limitations by applying recent advances in large-scale multi-task pre-training of generative Transformer models to the problem of VL-NLE tasks.
Our approach outperforms recent models by a large margin, with human annotators preferring the generated explanations over the ground-truth in two out of three evaluated datasets. As a novel challenge in VL-NLE research, we propose the problem of multi-task VL-NLE and show that jointly training on multiple tasks can increase the explanation quality. We discuss the ethical implications of high-quality NLE generation and other issues in recent VL-NLE research.
\end{abstract}

\section{Introduction}

Despite the enormous success of deep learning in numerous disciplines, widespread adoption of the technology is still inhibited by its lack of transparency: It is generally difficult to explain the decision-making process of a deep neural model, especially to the model's end-user. One promising approach is the generation of \emph{natural language explanations} (NLEs)~\cite{camburu_e-snli_2018}, which provide an intuitively understandable rationalization of a neural network's decision. Due to the growing need to integrate multi-modal information, such as vision and language, recent research has increasingly focused on generating \emph{natural language explanations for vision-language tasks} (VL-NLE) \cite{park_multimodal_2018, cho_unifying_2021,kayser_e-vil_2021,sammani_nlx-gpt_2022, majumder_knowledge-grounded_2022}. 

However, existing VL-NLE models suffer from at least one of the following issues: First, the models incorporate separate answer prediction and NLE generation modules (e.g., FME~\cite{park_multimodal_2018}, \eug~\cite{kayser_e-vil_2021}), which leads to inferior NLE generation performance due to the loose integration between the two modules. Second, their backbone models are often pretrained on a limited set of tasks (e.g., VL-T5~\cite{palaskar_advances_2022}), not fully exploiting the potential of multi-task learning with a unified architecture.  Third, they incorporate ad hoc solutions using additional task-specific modules or additional resources to increase their performance on the datasets at hand~(e.g., \nlxgpt\cite{sammani_nlx-gpt_2022}, RExC~\cite{majumder_knowledge-grounded_2022}), which does not allow them to be used as omnipotent models, a vision of current AI research~\cite{wang_ofa_2022,reed_generalist_2022}.

\begin{figure*}[ht]
    \centering
    \includegraphics[width=\textwidth]{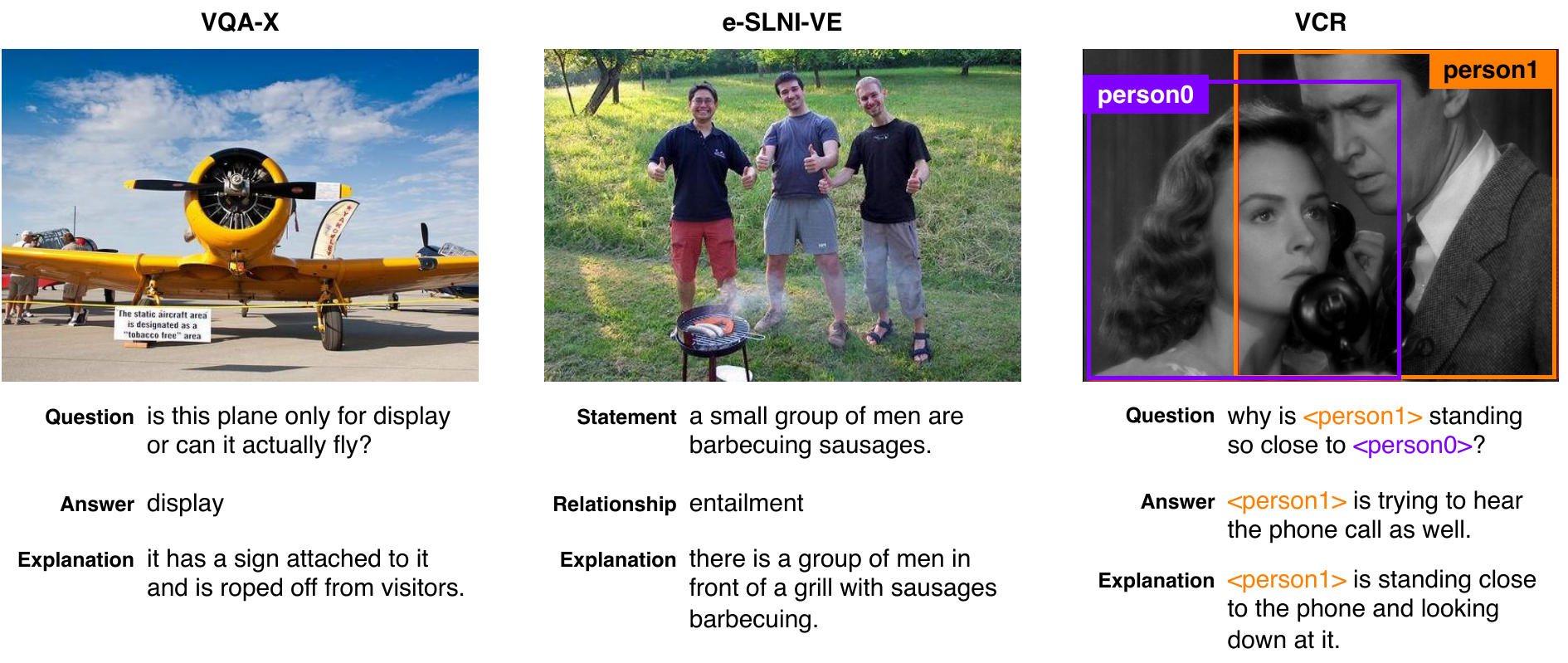}
    \caption{Example tasks from the datasets VQA-X~\cite{park_multimodal_2018}, e-SNLI-VE~\cite{kayser_e-vil_2021}, and VCR~\cite{zellers_recognition_2019}.}\label{fig:dataset_examples}
\end{figure*}

To overcome the mentioned issues, we propose to utilize recent advances in \emph{large-scale} \emph{multi-task} pre-training of generative vision-language models on VL-NLE tasks. Our assumption is that training on a broad range of diverse tasks enables learning the commonsense knowledge and reasoning capabilities necessary to generate complex and accurate NLEs. We fine-tune the recently presented OFA model~\cite{wang_ofa_2022}, a generative Transformer pretrained on a diverse set of multi-modal and uni-modal tasks, to each target NLE task. We show that a single unified architecture that is \emph{not} pretrained to generate explanations and \emph{without} any task-specific modifications and optimizations reaches state-of-the-art performance in the e-SNLI-VE and VQA-X datasets. The generated explanations received higher average ratings by human annotators than the ground-truth explanations on e-SNLI-VE while also being preferred in a direct A-B comparison on both e-SNLI-VE and VQA-X.
On the challenging VCR dataset, our approach outperforms other baselines according to our user study. As the explanation ratings achieved by our approach outperform those of the ground-truth annotations on two (i.e., e-SNLI-VE and VQA-X) of the three tasks, we call for new challenging VL-NLE tasks and improved dataset quality. As an immediate solution, we propose \mbox{\emph{e-ViL-combined}} as a new challenge, where a single model is required to solve and explain multiple different VL-NLE tasks from the three datasets. Evaluating our approach on VQA-X, e-SNLI-VE, and VCR after jointly training on the multi-VL-NLE dataset reveals its capability to achieve a competitive level of performance on e-SNLI-VE and VQA-X when compared to individually trained models, while considerably improving explanation quality on the VCR dataset.

\vspace{0cm}Our main contributions are as follows: First, we propose fine-tuning the OFA model~\cite{wang_ofa_2022} for VL-NLE tasks (\autoref{sec:approach}). Our approach outperforms the state-of-the-art baselines by a large margin in task accuracy and human explanation ratings, highlighting the importance of pre-training over architecture design. In several cases, our generated explanations are even preferred over the dataset ground-truth by the human participants of a user study (\autoref{sec:results}). We furthermore identify several issues in the current practice of evaluating VL-NLE models: For example, we identify the need for more challenging datasets and propose a new task, called e-ViL-combined (\autoref{sec:approach}). We also discuss the limitation of automatic explanation metrics (\autoref{sec:results}), and the necessity of metrics for measuring the faithfulness of explanations and undesired biases in the dataset (\autoref{sec:discussion}).

\section{Related Work}

\subsection{VL-NLE Datasets}

Motivated by the need to explain the behavior of vision-language (VL) models in human-understandable language, several datasets have been proposed recently. Illustrative example tasks for the three datasets described below can be found in~\autoref{fig:dataset_examples}.

The VQA-X dataset~\cite{park_multimodal_2018} is a subset of the VQA-V2 dataset~\cite{goyal_making_2017} that is annotated with human explanations. The images of the dataset are taken from the COCO dataset~\cite{lin_microsoft_2014}. Each task instance comprises a question and an accompanying image to answer it by predicting the correct answer class, where multiple answers can be correct.

The e-SNLI-VE dataset~\cite{kayser_e-vil_2021} was constructed by merging the e-SNLI~\cite{camburu_e-snli_2018} and the SNLI-VE~\cite{xie_visual_2019} datasets, which results in a visual entailment task with accompanying natural language explanations. In this task, a premise in the form of an image and a textual hypothesis are supplied. The task is to predict the relationship between the premise and the hypothesis as \textit{entailment}, \textit{contradiction}, or \textit{neutral}.

The Visual Commonsense Reasoning dataset (VCR)~\cite{zellers_recognition_2019} is a multiple choice (single answer) dataset of questions about images captured from movie scenes. Each instance has four answer possibilities that are appended to the input question. In the original VCR task formulation, a rationalization of the answer had to be selected from four possibilities as well, but the task was re-formulated as a free NLE-generation task for the e-ViL benchmark~\cite{kayser_e-vil_2021}.

The e-ViL benchmark, proposed by Kayser~\etal~\cite{kayser_e-vil_2021}, is an evaluation framework for models on vision-language and natural language explanation (VL-NLE) tasks. It consists of the three datasets VQA-X, e-SNLI-VE, and VCR, and provides two evaluation dimensions: (1) the VL-task label accuracy and (2) an explanation rating assessed by collecting the subjective ratings of human participants for the natural language explanations.

\subsection{VL-NLE Models}

Several VL-NLE models have been proposed in recent years to tackle the datasets mentioned above. They can be coarsely grouped into modular and unified models.

\paragraph{Modular Models.}%
The \eug model proposed by Kayser~\etal~\cite{kayser_e-vil_2021} is a modification of the Rationale\textsuperscript{VT} Transformer model proposed by Marasovic~\etal~\cite{marasovic_natural_2020}, who combined different vision-language models with a fine-tuned \gpt language model for rationale generation. Instead of the vision-language models proposed in \cite{marasovic_natural_2020}, \eug uses the UNITER vision-language Transformer as an answer prediction module to predict a task answer label. The representations used to predict the label, along with an additional answer and question text input, are then supplied to the \gpt explanation module to generate the rationale. Both \eug and Rationale\textsuperscript{VT} Transformer separate the NLE generation module from the answer prediction module, which necessitates heavier models with a high number of parameters and inferior performance due to the separation. A recent VL-NLE model RExC~\cite{majumder_knowledge-grounded_2022} adopts a complex modular approach: It first extracts bounding boxes of the relevant objects in the image with a VL model and then uses their features to generate knowledge snippets with a knowledge module. Finally, relevant knowledge snippets are fed to the NLE module and the resulting hidden vector to the prediction module. This highly modular approach by nature depends on several external models (i.e., Uniter~\cite{chen_uniter_2020} for bounding box generation, VisualCOMET~\cite{park_visualcomet_2020} for knowledge snippet generation and \gpt~\cite{brown_language_2020} for NLE generation) which requires a higher number of parameters as well as separate maintenance of the individual modules when applying the model to new data.

\paragraph{Unified Models.}%
\nlxgpt~\cite{sammani_nlx-gpt_2022} unifies answer prediction and explanation generation by training a distilled version~\cite{sanh_distilbert_2020} of \gpt Transformer decoder~\cite{brown_language_2020}. The decoder predicts a sequence of (question, answer, explanation) as the prediction target, and the encoded image is provided as the value and key vectors to the \gpt decoder's cross-attention layer. At inference time, the decoder takes as its input the question and autoregressively generates the answer and the explanation. \nlxgpt, however, is not trained on diverse tasks, such that it needs to resort to additional task-specific modules (e.g., a concept-detection module for e-SNLI-VE or bounding box projection-module for VCR) to improve its performance. VL-T5 and VL-BART are introduced in \cite{palaskar_advances_2022} and extend the Transformer-based unimodal language models BART-Base~\cite{lewis_bart_2020} and T5-Base~\cite{raffel_exploring_2020} with visual inputs (i.e., bounding box features from Faster R-CNN~\cite{ren_faster_2015}). Like \nlxgpt, the two models unify answer prediction and explanation generation. However, they also suffer from the same issue, i.e., they are not trained on diverse tasks, limiting task accuracy and explanation plausibility scores.

\subsection{Multi-task Learning with Unified Architectures}

Machine learning models trained on diverse tasks can profit from the interdependencies and transferability between tasks. The diversity of the tasks can, for example, provide a better inductive bias for a model and prevent it from overfitting to the training data~\cite{caruana_multitask_1997}. Also, simpler tasks can help the model to solve more complex tasks that are (partially) dependent on the simpler tasks in a curriculum learning manner~\cite{lee_what_2022}. In particular, multi-task learning datasets that involve multiple modalities can allow a model to benefit from the interaction between those modalities and solve general tasks and are used by recent pretrained Transformer-based multimodal architectures~\cite{kaiser_one_2017,pramanik_omninet_2020,lu_12--1_2020,hu_unit_2021,zhu_uni-perceiver_2022,reed_generalist_2022,wang_ofa_2022}. As such models are pretrained on large datasets and diverse tasks, they possess rich knowledge about the world. Furthermore, their ability to generate text allows them to present themselves as promising alternatives to hand-crafted and manually trained VL-NLE models. In this paper, we select the OFA model (``One for All'')~\cite{wang_ofa_2022}, which achieves state-of-the-art performance on multiple tasks, as the underlying architecture for our VL-NLE model.

\section{Approach}
\label{sec:approach}

The OFA model, proposed in early 2022 by Wang~\etal \cite{wang_ofa_2022}, is a recent vision-language model trained in a sequence-to-sequence multi-task learning setting. The authors employ large-scale  pre-training, allowing the model to learn robust vision and language representations across a variety of different tasks.
For completeness, we shortly summarize the original authors' approach below.

\subsection{Architecture}
\label{sec:architecture}

Architecture-wise, the OFA model is a standard encoder-decoder Transformer architecture as in~\cite{vaswani_transformer_2017, cho_encoder_decoder_2014}: All input modalities are processed by the same multi-headed self-attention encoder module. To accomplish this, both the image and input text are tokenized---using a ResNet-architecture as backbone~\cite{he_resnet_2016} for the tokenization of image patches and applying byte-pair encoding to the input text before its tokenization using a BERT-derived architecture~\cite{devlin_bert_2019}.

The OFA model is made effective by combining its vanilla Transformer architecture with a unified pre-training approach: The model is trained on a number of upstream tasks, such as image infilling, object detection, and text infilling. This allows the model to integrate a broad representation of implicit human knowledge into its downstream tasks. Out of the box, the provided pipeline supports fine-tuning for eight downstream tasks\footnote{The downstream tasks are visual grounding, grounded captioning, image captioning, image-text matching, visual question answering, image infilling, object detection, and text infilling.}---all of these share a common architecture and representation of input features. Every task's description is in purely textual form, no extra hints via specially constructed input features are given to the model. This makes the OFA model especially well suited to transfer its broad world knowledge to new tasks.

\subsection{Prompt Engineering for VL-NLE Tasks}
Following previous work on natural language explanations in vision-language settings~\cite{kayser_e-vil_2021, sammani_nlx-gpt_2022, palaskar_advances_2022}, we focus our experiments on the VQA-X \cite{park_multimodal_2018}, e-SNLI-VE \cite{kayser_e-vil_2021}, and VCR \cite{zellers_recognition_2019} datasets and the tasks associated with each. In addition, inspired by the OFA model's multi-task pre-training, we introduce a unified explanation task that combines the three datasets and tasks into one challenge designed to assess the model's multi-task performance. 

To follow the sequence-to-sequence requirement of the OFA architecture, we reformulate the datasets' task descriptions. As the OFA model does not utilize any specialized, task-specific methods, such as the inclusion of Faster R-CNN embeddings~\cite{kayser_e-vil_2021,marasovic_natural_2020} or bounding box embeddings~\cite{sammani_nlx-gpt_2022}, the target structure for each task is simple: take the image and text prompt as input and give the task-specific answer and corresponding explanation as natural language output. Similar to the approach of the \nlxgpt authors \cite{sammani_nlx-gpt_2022}, the same decoder module generates both the tasks' answers as well as the corresponding explanations. We refrain from utilizing a special separation token to distinguish answer and explanation but instead choose to follow the approach in \cite{sammani_nlx-gpt_2022} by simply using the word ``\textit{because}'' as a separator. Our approach reduces the task of designing task-specific modules to that of formulating appropriate prompts, following current trends in prompt engineering (c.f. \cite{liu2021pre}). In the following, we describe our task-specific prompt preparation steps.

\paragraph{VQA-X.} This task requires no further reformulation of the prompt due to its simplicity, but we diverge from the original OFA authors' approach for VQA \cite{antol_vqa_2015} by not constraining the output vocabulary and also not employing any complex decoding strategies, such as beam search, during inference. These choices increase inference speed and, from our preliminary testing, do not result in a notable difference in accuracy. We use this approach for all of our experiments.

\paragraph{e-SNLI-VE.} For training on e-SNLI-VE~\cite{kayser_e-vil_2021}, we follow the OFA authors' way of prompting tasks: The tasks' descriptions are changed to \textit{``Does the image describe {`[hypothesis]'}''}. This reformulation also necessarily changes the set of possible task answers from \textit{entailment, neutral}, and \textit{contradiction} to \textit{yes, maybe}, and \textit{no}, respectively.

\paragraph{VCR.} Past works have taken different approaches to the VCR \cite{zellers_recognition_2019} task. NLX-GPT \cite{sammani_nlx-gpt_2022} and VL-BART \cite{palaskar_advances_2022} omit the multiple choice aspect of the task and use the BERTScore as a proxy to decide if an answer is correct. \eug \cite{kayser_e-vil_2021} considers question-answer pairs individually and selects the final answer as the most likely pair. We choose to supply the answer possibilities by prompting the model with the question followed by the four different answers since VCR is designed as a multiple-choice task in which the correct answer must be chosen from four different options. This constrains our set of possible answers to be \textit{answer0} through \textit{answer3}. Furthermore, the questions and answers contain references to specific people or objects in the image, denoted by bounding boxes or regions. As the OFA model has already learned the representation of bounding boxes in textual form through the visual grounding pre-training \cite{wang_ofa_2022}, we follow that notation for denoting bounding boxes (see qualitative examples in Appendix). Since VCR does not include ground-truth explanations for the test set, we follow the dataset split proposed by Kayser~\etal \cite{kayser_e-vil_2021}.

\begin{table}[t]
\centering
\caption{Dataset statistics for VQA-X, e-SNLI-VE, VCR, and e-ViL-combined.}
\label{tab:statistics}
\begin{tabular}{lrrr}
  \toprule
  Dataset        & \# Train & \# Validation & \# Test \\
  \midrule
  VQA-X          & 29.5k    & 1.5k          & 2.0k    \\
  e-SNLI-VE      & 401.7k   & 14.3k         & 14.7k   \\
  VCR            & 212.9k   & 26.5k         & 25.2k   \\
  \midrule
  e-ViL-combined & 644.1k   & 42.3k         & 41.9k   \\
  \bottomrule
\end{tabular}%
\end{table}

\subsection{A Unified Vision-Language Explanation Task}
\label{subsec:evilcomb}
As a novel avenue for research in the field, we propose the multi-task VL-NLE dataset \mbox{\emph{e-ViL-combined}}, in which we train the model on all three tasks (i.e., VQA-X, e-SNLI-VE, VCR) of the e-ViL benchmark~\cite{kayser_e-vil_2021} simultaneously. With the multi-tasking approach, we hope to (i) enable research on NLE explanations for diverse tasks, such as investigating the facilitation of explanation generation across tasks, and (ii) offer a more challenging problem in the light of increasing VL-NLE model performance. \autoref{tab:statistics} shows the number of samples for each individual task and their respective fraction of the unified dataset. Since the datasets are very different in size, the tasks are represented in the resulting dataset in an unbalanced manner. However, as we evaluate the model performance on the \emph{individual} tasks, each dataset equally contributes to the overall model assessment. To evaluate our approach in this setting, we train OFA on the training splits of e-ViL-combined, which we henceforth refer to as \ofaxmt (``OFA-X multi-task'').

\section{Experiments}

\subsection{Fine-Tuning OFA}
We use the OFA authors' pretrained model weights for our experiments, which are available in different sizes on the project's GitHub page\footnote{\url{https://github.com/ofa-sys/ofa}}. The sizes we use for our experiments are: \textit{base} ($\sim$180M parameters), \textit{large} ($\sim$470M parameters), and \textit{huge} ($\sim$900M parameters). 

The model is fine-tuned on each task and studied for the effects of model size and pre-training tasks on both task accuracy and explanation quality.
We optimize using a cross-entropy loss with $0.1$ label smoothing and the Adam algorithm \cite{kingma_adam_2017}, employing a learning rate of $5e^{-5}$ with 1000 steps of warm-up and polynomial decay over 30000 training steps. The image and text embedding layer weights are kept frozen. For VQA-X, VCR, e-SNLI-VE, and the unified task, we use effective batch sizes of 16, 64, 128, and 128, respectively. The training process of the \textit{``large''} model variants takes approximately 24 hours on a single NVIDIA A100 GPU. Please refer to the attached training code within the supplementary material to find out about our exact hyperparameter settings.

\subsection{Evaluation Metrics}
VL-NLE models are evaluated with respect to three main aspects: First, the model's prediction performance on the primary task. Second, the model's explanation quality is measured by automatic metrics, which compare the generated explanations against the ground-truth explanations. Third, the evaluation of the model's explanation generation performance by human participants as part of a user study.

\begin{table*}[t]
  \centering
  \caption{Explanation and task scores for VQA-X, e-SNLI-VE, and VCR. Presented metrics are Meteor (M), BERTScore (BS), human evaluation (Human), and task accuracy (Acc.). We also provide the ground-truth explanation scores from our human evaluation.}
  \begin{minipage}[t]{1.0\linewidth}
    \centering
    \label{tab:results}
    \begin{tabular}{lcccc@{\hskip 2em}cccc@{\hskip 2em}cccc}
      \toprule
      & \multicolumn{4}{c}{VQA-X} & \multicolumn{4}{c}{e-SNLI-VE} & \multicolumn{4}{c}{VCR}                                                                                                                                       \\
      \cmidrule(lr){2-5}\cmidrule(lr){6-9}\cmidrule(lr){10-13}
                                                     & M                         & BS                            & Human         & Acc.          & M             & BS            & Human         & Acc.          & M             & BS            & Human         & Acc.          \\
      \midrule
      Ground-truth                                      & -                         &      -                       & 89.9          &   -           &   -            & -             & 85.0          & -             &  -            & -             & 87.6         &  -            \\ \midrule
      PJ-X \cite{park_multimodal_2018}               & 19.7                      & 84.6                          & 65.4          & 76.4          & 14.7          & 79.1          & 59.6          & 69.2          & 20.5          & 78.4          & 52.7          & 39.0          \\
      FME \cite{wu_faithful_2019}                    & 20.4                      & 85.2                          & 63.2          & 75.5          & 15.6          & 79.7          & 58.5          & 73.7          & 22.7          & 79.4          & 58.5          & 48.9          \\
      \eug \cite{kayser_e-vil_2021}                  & 22.1                      & 87.0                          & 71.5          & 80.5          & \textbf{19.6} & \textbf{81.7} & 68.9          & 79.5          & 22.5          & 79.0          & 65.1          & 69.8          \\
      \nlxgpt \cite{sammani_nlx-gpt_2022}            & 23.1                      & 86.9                          & 83.2          & 83.0          & 18.8          & 80.8          & 67.4          & 73.91         & \textbf{26.4} & \textbf{80.3} & -             & -             \\
      \midrule
      OFA-X                                          & \textbf{24.5}             & \textbf{87.2}                 & \textbf{89.5} & 91.2          & 18.6          & 80.9          & \textbf{85.7} & \textbf{80.9} & 12.2         & 79.8          & 68.9          & \textbf{71.2} \\
      OFA-X\textsubscript{\textsf{MT}}                         & 23.1                      & 86.8                          & 87.8          & \textbf{92.6} & 17.9          & 80.3          & 80.4         & 78.9         & 11.3         & 79.3          & \textbf{77.3} & 62.0          \\
      \bottomrule
    \end{tabular}
  \end{minipage}
\end{table*}

\paragraph{Task evaluation.}
In the VQA-X dataset, there are usually multiple correct options per task, which each award a score between $\frac{1}{3}$ and $1$. The other two tasks e-SNLI-VE and VCR award a score of $1$ for the single correct answer and $0$ otherwise. Good performance on these scores is crucial, as it measures the model's ability to correctly solve the task---a core requirement for a meaningful explanation. For our user study, we follow previous approaches~\cite{kayser_e-vil_2021, sammani_nlx-gpt_2022, palaskar_advances_2022} and only evaluate model explanations for which the task was answered correctly.

\paragraph{Automatic evaluation.}
Automatic explanation evaluation metrics are often used for inexpensive comparisons between models, {e.g.,} for hyperparameter optimizations. 
This inexpensive evaluation comes at a cost: The metrics only compare semantic similarity between the prediction and ground-truth explanations and cannot take image information into account or evaluate the correctness or sufficiency of the explanation. Their appropriateness for use with VL-NLE tasks is further discussed in \autoref{sec:expl_results}. We use the BERTScore \cite{Zhang2020BERTScoreET} as the primary automatic metric for explanation validation as Kayser~\etal \cite{kayser_e-vil_2021} find that it has the highest correlation to human judgment. More specifically, we follow \cite{kayser_e-vil_2021} in using the \texttt{distilbert-base-uncased} model from the HuggingFace Datasets library~\cite{LhoestHuggingfaceDatasets}, which is based on DistilBERT~\cite{sanh_distilbert_2020}. We follow common convention and evaluate the explanations using multiple natural language generation metrics, such as BLEU \cite{Papineni2002BleuAM}, METEOR \cite{Banerjee2005METEORAA}, ROUGE-L \cite{Lin2004ROUGEAP}, CIDEr \cite{Vedantam2015CIDErCI}, and SPICE \cite{Anderson2016SPICESP}, visible in the full results table in Appendix.
Also, for all automatic metrics, we report the ``\textit{filtered}'' scores, by only counting explanations for which the model's corresponding answer was correct. For reference, we also provide and elaborate on the \textit{``unfiltered''} and \textit{``scaled''} model evaluations in Appendix. 

\paragraph{Human evaluation.}
In addition to the automatic evaluation, we perform an extensive user study with \emph{Amazon Mechanical Turk} (MTurk), following the framework proposed by Kayser~\etal~\cite{kayser_e-vil_2021}: Participants are first asked to select the task answer, then to individually rate both the ground-truth and model explanation on the question: ``\textit{Does the explanation justify the answer}''. The answer options are ``\textit{yes}'', ``\textit{weak yes}'', ``\textit{weak no}'', and ``\textit{no}'' which are mapped from 1 to 0 for the rating. When survey participants rate an explanation with ``\textit{weak no}'' or ``\textit{no}'', they are prompted to justify their decision. They can choose one or multiple of ``\textit{confusing sentence}'', ``\textit{insufficient justification}'', or ``\textit{incorrect description of image}''. 
By asking participants for the known and validated task answer, we can effectively exclude participants who do not properly understand the dataset task at hand in our statistical evaluation later on.

For every tested model, we select 300 random image/question combinations from the respective test set, which the trained model correctly solved. Every task is then completed by at least five different human annotators. In line with \cite{kayser_e-vil_2021}, we exclude all samples from the evaluation where the task was not answered correctly. To qualify for our user study, the annotators had to have a minimum 90\% HIT acceptance rating and possess the MTurk master's qualification. Because our evaluation is based on \cite{kayser_e-vil_2021}, it is also in line with those of competing approaches \cite{sammani_nlx-gpt_2022, palaskar_advances_2022}.\footnote{Different from the proposed framework by Kayser~\etal~\cite{kayser_e-vil_2021} we do not perform browser checks on the validity of the given answers by our participants: Kayser~\etal required at least 80\% of the task answers to be correct for an MTurk worker assignment to be submit-able. We argue that by not hinting participants at the correct answers, our evaluation ensures that only those explanations are counted, for which the task was understood correctly by the human annotator.}
In addition to the common model evaluation questions, we ask the participants which explanation they prefer or whether their quality is perceived as equal.

We use the \textit{large} model variant (470M parameters) instead of the \textit{huge} model (900M) parameters for our human evaluation to have a model size more comparable to other works. This stresses the point that our improved evaluation results are not primarily caused by an increase in model size. We use the BERTScore to select which model flavor to evaluate for a given task. 

\subsection{Baselines}
For our evaluation, we select the state-of-the-art baselines from \cite{sammani_nlx-gpt_2022}, which are PJ-X~\cite{park_multimodal_2018}, FME~\cite{wu_faithful_2019}, e-UG~\cite{kayser_e-vil_2021}, and NLX-GPT~\cite{sammani_nlx-gpt_2022}. We exclude VL-BART, VL-T5 \cite{palaskar_advances_2022}, and RExC \cite{majumder_knowledge-grounded_2022} from our analysis as they can currently not be verified due to missing source-code.\footnote{Cf.~\url{https://github.com/majumderb/rexc/issues/1}.} We include their reported results in Appendix.

\section{Results}
\label{sec:results}
To evaluate our approach, we first report task accuracy and human explanation ratings in accordance with previous works in the field. We present detailed results on the quality of explanations and compare them with the ground-truth explanations (qualitative examples in Appendix). To study the effect of different model sizes and pre-training tasks on model performance, we present an ablation study. Finally, we compare the individually trained models with the \ofaxmt model trained on the e-ViL-combined multi-tasking dataset (cf.~\autoref{subsec:evilcomb}).

\subsection{Task Accuracy and Explanation Rating}
\label{sec:expl_results}
On each of the three tasks in the e-ViL benchmark~\cite{kayser_e-vil_2021}, our approach significantly outperforms all publicly available baselines on both task accuracy and the average explanation quality rating by human annotators. \autoref{tab:results} shows the results of our best models on each task. Additionally, we present the results of the \ofaxmt model trained on the proposed e-ViL-combined dataset in a separate row. We deliberately choose to include both the BERTScore and the METEOR score, since they were previously identified as the metrics with the highest correlation to human rating decisions \cite{kayser_e-vil_2021}. Somewhat counterintuitive, considering the quality of our results in human ratings, our approach does not perform better on all automatic metrics and tasks. This reinforces the findings by Kayser~\etal \cite{kayser_e-vil_2021} that automatic metrics alone are not fully indicative of actual explanation quality.

The \ofaxmt model, trained on the e-ViL-combined task, performs similarly well when compared to the individually fine-tuned models. On the human evaluation of explanations on the VCR dataset, the unified model achieves significantly higher scores when comparing it to its single-task counterpart, thus demonstrating that training on a more diverse dataset can improve explanation quality. Moreover, the \ofaxmt also outperforms its VQA-X counterpart on task accuracy. 
These findings further illustrate the benefits of multi-task pre-training and the flexibility of the sequence-to-sequence learning approach for VL-NLE tasks.
See the Appendix for a full comparison table using all common metrics.

\subsection{Explanation Quality and Comparison to Ground-Truth}
\label{sec:groundtruth}
When comparing to the ground-truth annotations, we observe that our approach receives higher explanation ratings on the e-SNLI-VE dataset and only marginally worse ratings on VQA-X. While the comparison of human preference between ground-truth and OFA-X explanations, visible in \autoref{tab:preference}, shows a preference for the ground-truth explanations in 53.4\% of the cases on VCR, participants on average preferred our explanation over the ground-truth on both the VQA-X and e-SNLI-VE datasets.

\autoref{tab:downsides} shows a comparison of explanation shortcomings of our results and other competing approaches for which this data is publicly available. Note that this is the average of all three tasks and the percentage of total explanations that were marked with these shortcomings. Based on these metrics, we can deduce that significantly fewer of our explanations had common flaws than reported by other works. In addition, we find that confusing or nonsensical explanations are much less common with our approach.

\begin{table}[t]
\centering
\caption{Comparison of human preference between ground-truth (GT) explanation and OFA-X explanations.}
\label{tab:preference}
\begin{tabular}{@{}cccc@{}}
\toprule
          & Prefer ours   & No preference & Prefer GT     \\ \midrule
VQA-X     & \textbf{43.9\%} & 25.0\%         & 31.1\%          \\
e-SNLI-VE & \textbf{41.1\%} & 24.4\%         & 34.0\%        \\
VCR       & 26.4\%          &  20.2\%         & \textbf{53.4\%} \\ \bottomrule
\end{tabular}%
\end{table}

\begin{table}[t]
\centering
\caption{Evaluation of explanation shortcomings. Numbers are in \% of total rated samples.}
\label{tab:downsides}
\begin{tabular}{@{}cccc@{}}
\toprule
      & Untrue to & Lack of & Non-sensical \\ 
      & Image     & Justification & Sentence \\\midrule
PJ-X \cite{park_multimodal_2018}  & 25.0\%            & 26.4\%                  & 8.9\%                   \\
FME \cite{wu_faithful_2019}   & 21.8\%            & 23.1\%                  & 13.7\%                  \\
\eug \cite{kayser_e-vil_2021} & 15.9\%            & 25.0\%                  & 7.4\%                   \\
OFA-X & \textbf{10.8\%}   & \textbf{17.3\%}         & \textbf{3.4\%}         \\ \bottomrule
\end{tabular}%
\end{table}

\subsection{Model Size and Pre-training Ablations}
In this section, we study the effect of three different model sizes and pre-training methods on model performance: For each of the tasks, we evaluate multiple different methods;  {\autoref{tab:ablations}} summarizes our findings. In the pre-training column, \textit{OFA} refers to the pretrained model checkpoint for the specific size release by Wang~\etal \cite{wang_ofa_2022}. As they also perform fine-tuning on multiple downstream tasks, including image captioning on MSCOCO captions \cite{lin_microsoft_2014} and visual entailment on SNLI-VE \cite{xie_visual_2019}. Sammani~\etal \cite{sammani_nlx-gpt_2022} uses image captioning as a pre-training task for NLX-GPT. Due to the descriptive nature of the image captioning task, we agree that it is a suitable baseline for VL-NLE models. As a starting point for our training, we, therefore, experiment with using the checkpoint provided by Wang~\etal \cite{wang_ofa_2022} that is fine-tuned on the downstream image captioning task. We observe that, while this approach results in higher scores on the automatic metrics compared to the general OFA pre-training on the VQA-X dataset, it does not translate to the other tasks. We use the same approach, starting from the checkpoint fine-tuned on SNLI-VE, for e-SNLI-VE but find that while the task score is slightly higher, the explanation scores are equally reduced.

Across the board, we observe the \textit{huge} model variants outperforming the other model sizes' task performance: This is most notable on the VCR task with a 3.7\% absolute increase in prediction accuracy. We hypothesize that the larger model size aids in understanding the more complex task. In terms of automatic metrics, the \textit{large} models consistently achieve a higher BERTScore than the corresponding huge model. 

To test whether the task accuracy improvements of the \textit{huge} models translate to improved explanation quality, we perform an additional human evaluation on the \textit{huge} model trained on the VCR dataset, since our models perform worst on this task. We also observe the largest task score improvement between \textit{large} and \textit{huge} here as well. Our findings further support the argument that the BERTScore may indicate better explanations but can also be misleading. The \textit{huge} model scores significantly higher than its \textit{large} counterpart, with a human explanation rating of 77.8\% compared 68.9\%, indicating that the explanations of the other \textit{huge} models may also be better. We leave this question open for future research since the scores for the \textit{large} models are sufficiently high and also more comparable to the model sizes of previous works.

After fine-tuning \ofaxmt on the \textit{e-ViL-combined} dataset, the model performs comparatively well regarding the {BERTScore} when compared to the models trained on the singular tasks; for VQA-X it even reaches a higher task accuracy. The only significant accuracy decrease is on the VCR task, where apparently more specialized training is required. We evaluate the quality of the explanations with another human evaluation visible in the last row of \autoref{tab:results}. We find that the quality of explanations is on par with or exceeding, in the case of VCR, the models trained only on one task. These findings support the claims by Wang~\etal \cite{wang_ofa_2022} that multi-task training is beneficial to a model's reasoning abilities.

Finally, to assess whether the additional task of generating explanations effects task accuracy, we train an identical baseline model on each individual task that only outputs task answers. We observe the task accuracy to be comparable to the VL-NLE models on VQA-X and e-SNLI-VE. For VCR, generating explanations seems to aid the model during training, leading to a higher task score than the non-explanation baseline.

\begin{table}[t]
  \footnotesize%
  \centering
  \caption{Studies on the effect of model size and pre-training on model performance. \textit{``w/o expl''} refers to models trained on just the main task, omitting explanation generation. We also present the scores for the model trained on \textit{e-ViL-combined} for the individual datasets and the combined dataset. The models with the highest BERTScore (BS) are used for human evaluation.}
  \label{tab:ablations}
  \begin{tabular}{cccccc}
    \toprule
    Test                     & Train                    & Size  & Pretrain. & BS            & Acc.          \\ \midrule
    \multirow{5}*{VQA-X}{}   & \multirow{5}*{VQA-X}{}   & Base  & OFA       & 86.1          & 90.0          \\
                             &                          & Large & OFA       & 86.8          & 92.3          \\
                             &                          & Large & Caption   & \textbf{87.2} & 91.2          \\
                             &                          & Huge  & OFA       & 86.7          & 93.9          \\
                             &                          & Huge  & Caption   & 86.9          & \textbf{94.0} \\
    (w/o expl.)              & (w/o expl.)              & Large & OFA       & -             & 93.1          \\
    VQA-X                    & e-ViL-comb.              & Large & OFA       & 86.8          & 92.6          \\ \midrule
    \multirow{5}*{e-SNLI-VE} & \multirow{5}*{e-SNLI-VE} & Base  & OFA       & 80.6          & 77.6          \\
                             &                          & Large & OFA       & \textbf{80.9} & 80.9          \\
                             &                          & Large & Caption   & 80.9          & 81.0          \\
                             &                          & Large & SNLI-VE   & 80.3          & 81.5          \\
                             &                          & Huge  & OFA       & 80.8          & \textbf{82.1} \\
    (w/o expl.)              & (w/o expl.)              & Large & OFA       & -             & 81.4          \\
    e-SNLI-VE                & e-ViL-comb.              & Large & OFA       & 80.4          & 78.9          \\ \midrule
    \multirow{4}*{VCR}       & \multirow{4}*{VCR}       & Base  & OFA       & 79.5          & 67.5          \\
                             &                          & Large & OFA       & \textbf{79.8} & 71.2          \\
                             &                          & Large & Caption   & 79.5          & 67.2          \\
                             &                          & Huge  & OFA       & 79.5          & \textbf{74.9} \\
    (w/o expl.)              & (w/o expl.)              & Large & OFA       & -             & 70.3          \\
    VCR                      & e-ViL-comb.              & Large & OFA       & 79.3          & 62.0          \\ \midrule
    e-ViL-comb.              & e-ViL-comb.              & Large & OFA       & 82.2          & 77.8          \\ \bottomrule
  \end{tabular}%
\end{table}

\section{Discussion}
\label{sec:discussion}
Within our user study, our approach yields human explanation ratings that exceed or match the ground-truth annotation quality in two of the three evaluated datasets. Both OFA-X and the multi-tasking variant \ofaxmt outperform all publicly available baselines. These results are based on the OFA model, which is closely following the reference Transformer encoder-decoder architecture. This indicates that---for VL-NLE tasks---choosing the right pre-training strategy impacts model performance more strongly than architecture.

The achieved explanation ratings, which are above the ground-truth level, call attention to the need for new datasets with increased annotation quality, and, ultimately, new directions for VL-NLE research. With \emph{e-ViL-combined}, we are proposing the diversification of VL-NLE datasets to multi-tasking scenarios and suggest further widening the scope of possible tasks. Unified sequence-to-sequence models such as OFA enable the possibility of explanation generation to a wide variety of tasks, even including so far disregarded tasks such as image generation or captioning. We believe that through multi-task learning, the generation of high-quality NLEs on individually challenging datasets can be facilitated, as illustrated by the improved results on the VCR dataset.

\paragraph{Ethical concerns.} Current evaluation strategy of using human explanation ratings collected in the usual VL-NLE evaluations only assesses a particular aspect of the explanation quality, namely their \textit{plausiblity} or \textit{persuasiveness}~\cite{jacovi_towards_2020}. However, the favorable ratings of the explanations, and their preference above ground-truth annotations in many cases, call attention to the measurement of explanation \textit{faithfulness}, referring to the degree to which a generated explanation truly represents the reasoning process of a model, as opposed to a pure rationalization that is not reflecting the process of the answer prediction. 
How to measure the faithfulness of an explanation is an actively discussed and challenging issue~\cite{jacovi_towards_2020, deyoung_eraser_2020, wiegreffe_measuring_2021}. For example, Wiegreffe~\etal proposed two evaluation methods to assess necessary conditions that need to be fulfilled for faithful explanations~\cite{wiegreffe_measuring_2021}. However, they currently do not allow for a comparative analysis of different VL-NLE models and, therefore, provide no information about the degree of a particular model's faithfulness. With the increasing persuasiveness of the generated explanations, ethical issues of deploying these models without knowledge about the explanation faithfulness become decidedly clear. Reinforcing statements from recent publications~\cite[Section 5]{majumder_knowledge-grounded_2022}, we argue that the development of faithfulness metrics should be prioritized by the community.

Finally, we would like to highlight ethical concerns with the current VL-NLE datasets: As the human annotations used to fine-tune our models contain several undesirable biases, such as racial and gender stereotypes, the model is also trained to reproduce some of these potentially harmful concepts. We highlight a few of these examples within our Appendix. Naturally, due to the descriptive way of formulation for VL-NLE tasks, image subjects are generally referred to by their external attributes. This superficial and incomplete representation of subjects quickly leads to the inclusion of common societal stereotypes by human annotators while creating the dataset. While the inappropriate representation of external human attributes like gender and/or ethnicity can be harmful to affected individuals, we do not regard this possibility as a reason to hold back on the publication of our research. Instead, to make the research community aware of this issue, we choose to transparently communicate it. For the creation of widely employable AI, discrimination that could result from improper or incautious model use must be effectively prevented: We see a clear need for more efforts to reduce the potentially harmful bias in such datasets, by introducing dedicated quality assurance measures (cf.~\cite{mehrabi_survey_2021,schwartz_towards_2022}).

\section{Conclusion}

We leverage the diverse multi-task pre-training of the OFA model to fine-tune it on three common VL-NLE tasks and achieve state-of-the-art results on each. As the presented approach is generally preferred over the ground-truth explanations in two of three evaluated datasets, we notice the need for improved and extended VL-NLE datasets. As an immediate solution, we propose \textit{e-ViL-combined}, a multi-task dataset for VL-NLE models. We showed that training the OFA model jointly on this diverse dataset can improve explanation quality in the case of the VCR dataset. We argue that diverse multi-task training---in contrast to architectural modifications---is key to achieving higher-quality NLEs. Finally, we emphasize two important open issues in VL-NLE research: (i) The development of comparative NLE faithfulness metrics, and (ii) undesired biases in current VL-NLE datasets.

\paragraph{Acknowledgements.}
The authors gratefully acknowledge support from the Federal Ministry of Education and Research 
(BMBF), the Free and Hanseatic City of Hamburg under the Excellence Strategy of the Federal Government and the Länder, and the German Research Foundation DFG for the projects CML TRR169, MoReSpace and LeCareBot.

{%
  \small%
  \bibliographystyle{ieee_fullname}%
  \bibliography{bibliography}%
}

\clearpage
\appendix

\section{Appendix}
\label{sec:appendix}
\subsection{VCR Bounding Box Notation}

\begin{figure}[h]
    \centering
    \includegraphics[width=.9\columnwidth]{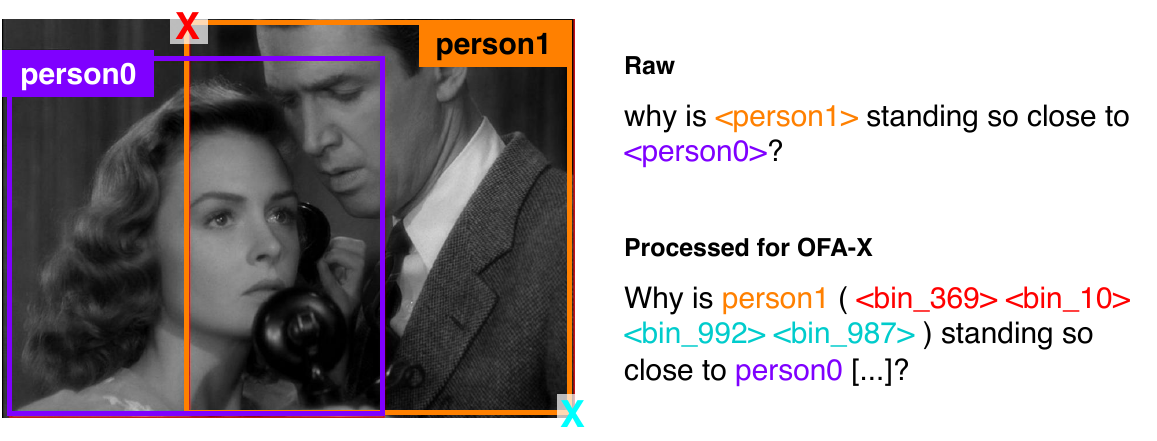}
    \caption{Example of how bounding boxes within questions from the VCR dataset~\cite{zellers_recognition_2019} are passed to the OFA-X model.}\label{fig:bounding_boxes}
\end{figure}

As part of the pre-training procedure~\cite{wang_ofa_2022}, the OFA model is trained on recognizing a textual notation for bounding boxes. To this end, the input images are quantized into 1000 bins in the X and Y dimension. The top-left and bottom-right corners of the bounding box are then encoded into special text tokens representing the individual coordinates. For the VCR task, we concatenate the bounding box notation of a person to their respective text token. This approach avoids the need to design separate and specific bounding box input features, as for example required by NLX-GPT~\cite{sammani_nlx-gpt_2022}. To ensure that we do not artificially inflate the scores on the automated metrics, we remove these bounding boxes from our generated explanations and the corresponding ground-truth explanations before evaluation. An example of the notation used is illustrated in \autoref{fig:bounding_boxes}.

\subsection{Qualitative examples}

\begin{figure*}[h]
    \centering
    \includegraphics[width=.85\textwidth]{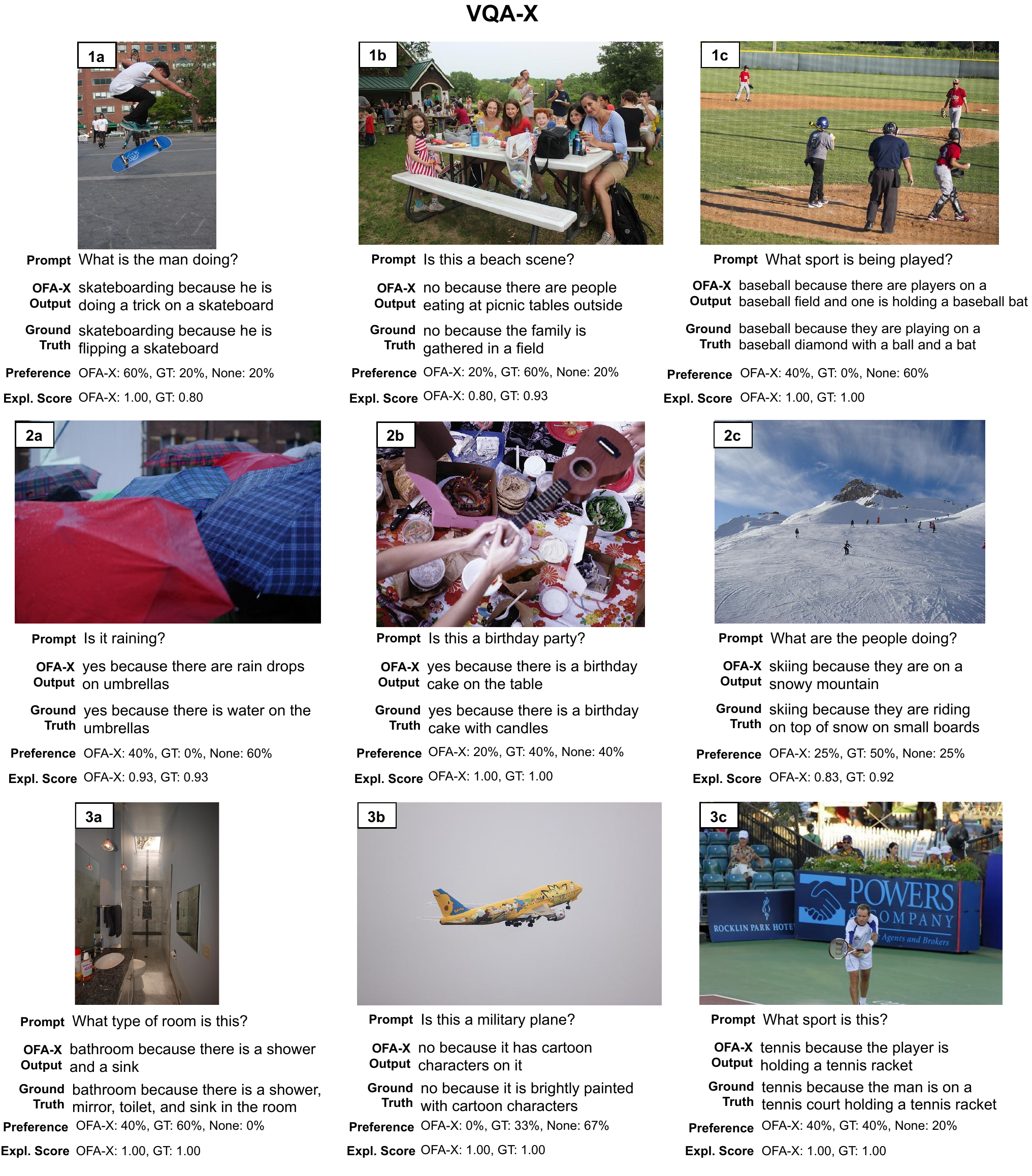}
    \caption{Qualitative comparison of OFA-X's outputs and ground-truths for the VQA-X dataset~\cite{antol_vqa_2015}. Also includes preference and explanation ratings from the human evaluation.}\label{fig:qualitative_vqax}
\end{figure*}

\begin{figure*}[h]
    \centering
    \includegraphics[width=.85\textwidth]{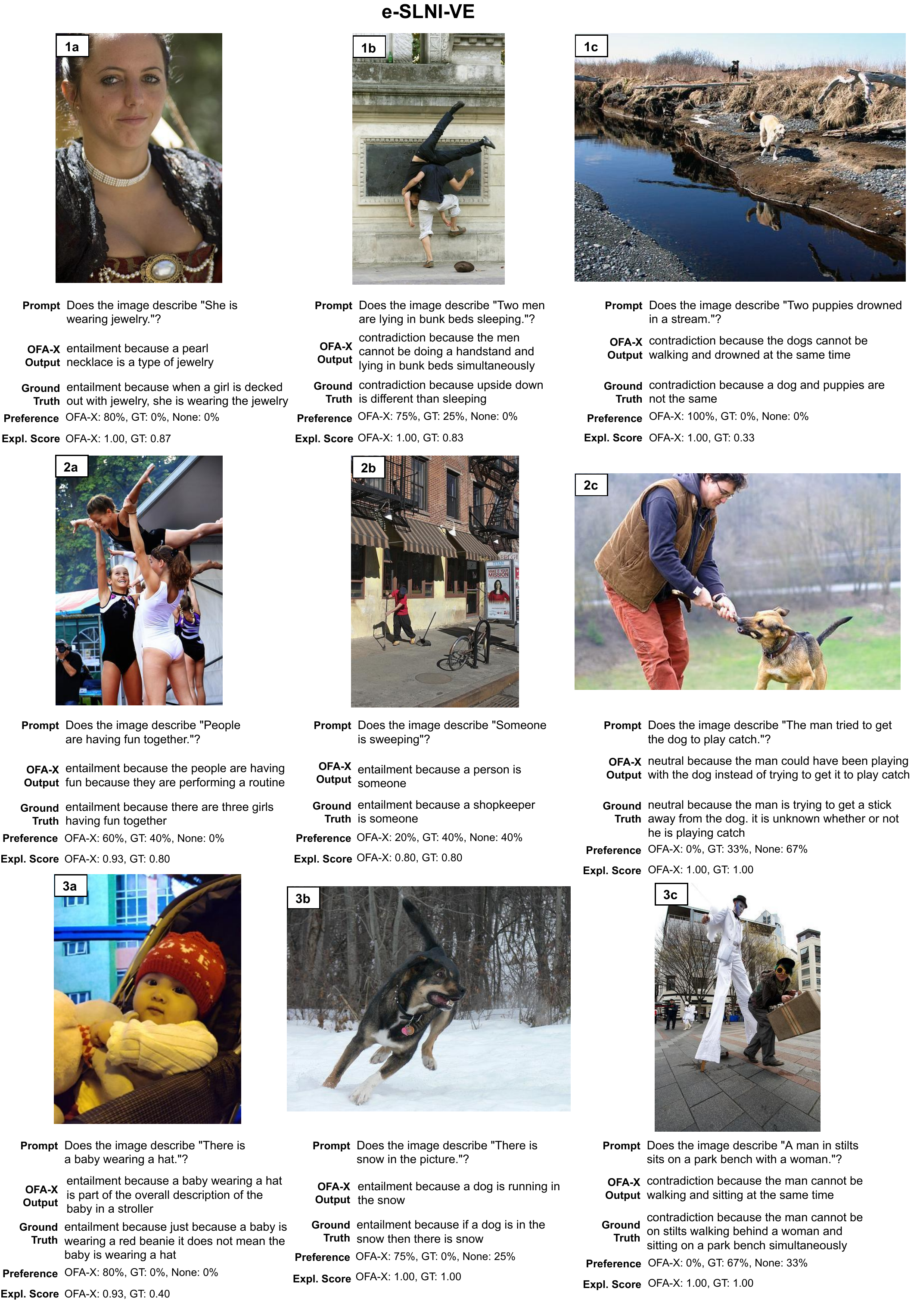}
    \caption{Qualitative comparison of OFA-X's outputs and ground-truths for the e-SNLI-VE dataset~\cite{kayser_e-vil_2021}. Also includes preference and explanation ratings from the human evaluation.}\label{fig:qualitative_esnlive}
\end{figure*}

\begin{figure*}[h]
    \centering
    \includegraphics[width=.85\textwidth]{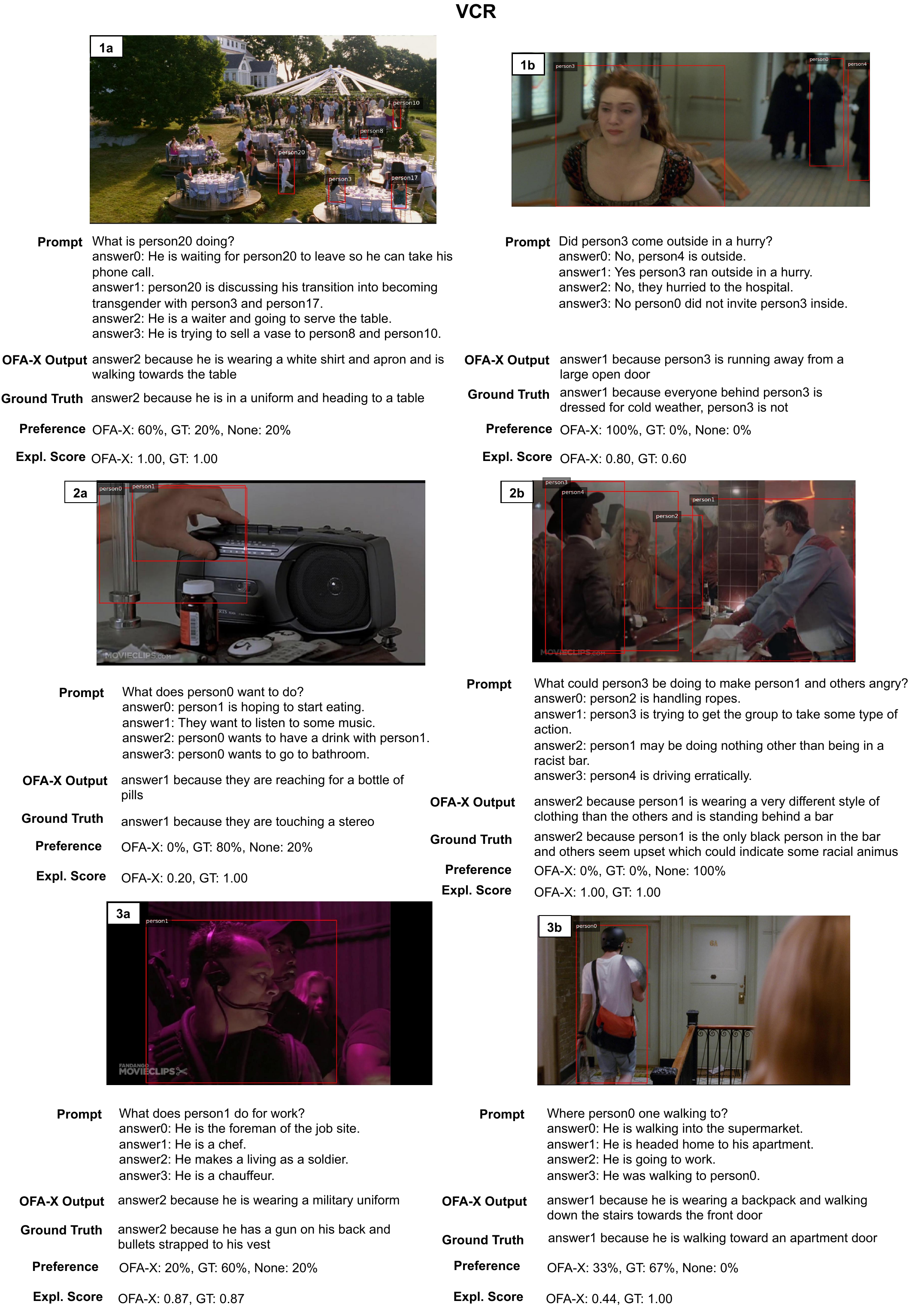}
    \caption{Qualitative comparison of OFA-X's outputs and ground-truths for the VCR dataset~\cite{zellers_recognition_2019}. Due to space constraints, the bounding box information within the prompts has been removed: The \texttt{personN} tokens have to be exchanged for the notation as described in \autoref{fig:bounding_boxes}. The visual bounding boxes are not part of the prompt–––they were added as a visual cue for the human evaluators.
    Also includes preference and explanation ratings from the human evaluation.}\label{fig:qualitative_vcr}
\end{figure*}

We provide some qualitative examples that are correctly solved by the model. They are not hand-picked but are chosen from the first examples from the user study for each task, so that they show a realistic picture of the model's performance. Images are cropped to fit into the grid layout. See \autoref{fig:qualitative_vqax}, \autoref{fig:qualitative_esnlive}, and \autoref{fig:qualitative_vcr}.

\subsection{OFA-X vs NLX-GPT}
We qualitatively compare OFA-X against the best-performing baseline NLX-GPT on the first 27 examples of the VQA-X dataset (see \autoref{tab:ofaxvsnlxgpt1}, \autoref{tab:ofaxvsnlxgpt2}, and \autoref{tab:ofaxvsnlxgpt3}). To this end, we use results from the same OFA-X model that is used for human evaluation (i.e., OFA-X \textit{large} pretrained on the MSCOCO caption dataset) and the results from NLX-GPT provided by the authors on their GitHub page\footnote{\url{https://github.com/fawazsammani/nlxgpt}}. We observe that OFA-X generates overall more correct answers (examples \#1, \#2, \#4, \#5, \#7, \#11, \#13), better handles the perspective of objects (examples \#1 and \#13), gives more plausible and concrete explanations (examples \#0, \#9, \#12, \#14, \#16, \#17, \#21, \#22). On the other hand, NLX-GPT is inferior in all the mentioned aspects and its explanations are even contradictory (examples \#7, \#11, \#19).

\subsection{Detailed results}

In \autoref{tab:full-results}, we present detailed results of OFA-X and \ofaxmt on automatic explanation metrics employed by previous results. We compare our results with previous works, including those reported by Paleskar \etal~\cite{palaskar_advances_2022} and Majumder \etal~\cite{majumder_knowledge-grounded_2022}, where public source code is currently not available to verify the scores reported. 

\subsection{Filtered, Unfiltered and Scaled Explanation Metrics}

Previous works have presented the automatic explanation metric results of their VL-NLE models in different ways. For example, Sammani \etal~\cite{sammani_nlx-gpt_2022} note that some works evaluated explanation metrics only if the corresponding task was correctly answered (\textit{filtered}), while others evaluated all explanations (\textit{unfiltered}). As a third method, the automatic explanation metrics are \textit{scaled} by the task score, e.g. 1.0 for a correct answer or 0 for an incorrect answer. The scaled method was described for example by Paleskar \etal~\cite{palaskar_advances_2022}. However, their presented results---to the best of our knowledge---correspond to the filtered method. To compare the effect on the metrics when using these different approaches, we present \textit{filtered, unfiltered} and \textit{scaled} automatic explanation metrics results in \autoref{tab:fitlered-unfiltrd-scaled}.

\subsection{OFA-X Pre-training}
\label{sec:ofa-x-pre}

As OFA-X is based on the pretrained OFA, it also shares OFA's pre-training dataset, where OFA avoids data leakage by excluding from the training set the images that appear in the validation and test sets of the downstream tasks (cf.~\cite[Appendix A.1]{wang_ofa_2022}).\footnote{The datasets for OFA pre-training can be found here: \url{https://github.com/OFA-Sys/OFA/blob/main/datasets.md}} In the following we describe why OFA-X is free from leakage of its downstream test datasets.
\begin{itemize}
\item VQA-X is based on the VQA and VQA v2.0 datasets~\cite{antol_vqa_2015,goyal_making_2017}, where VQA v2.0 is an extension of the VQA dataset with additional complementary images to balance the VQA dataset. As VQA v2.0 is used for fine-tuning OFA, OFA-X is not pretrained on VQA v2.0, and accordingly, there is no leakage of VQA-X.

\item eSNLI-VE is an integration of the explanations from e-SNLI~\cite{camburu_e-snli_2018} with the image-sentence pairs from SNLI-VE~\cite{xie_visual_2019}. Since OFA is fine-tuned on SNLI-VE, there is no leakage of eSNLI-VE.

\item VCR is based on images from Large Scale Movie Description Challenge (they are from Blu-ray movies)~\cite{rohrbach_movie_2017} and images on YouTube videos~\cite{zellers_recognition_2019}. OFA, and thus OFA-X, is not pretrained on these datasets.
\end{itemize}

\subsection{Dataset Biases and Ethical Concerns}

\begin{figure*}[t]
    \centering
    \includegraphics[width=\textwidth]{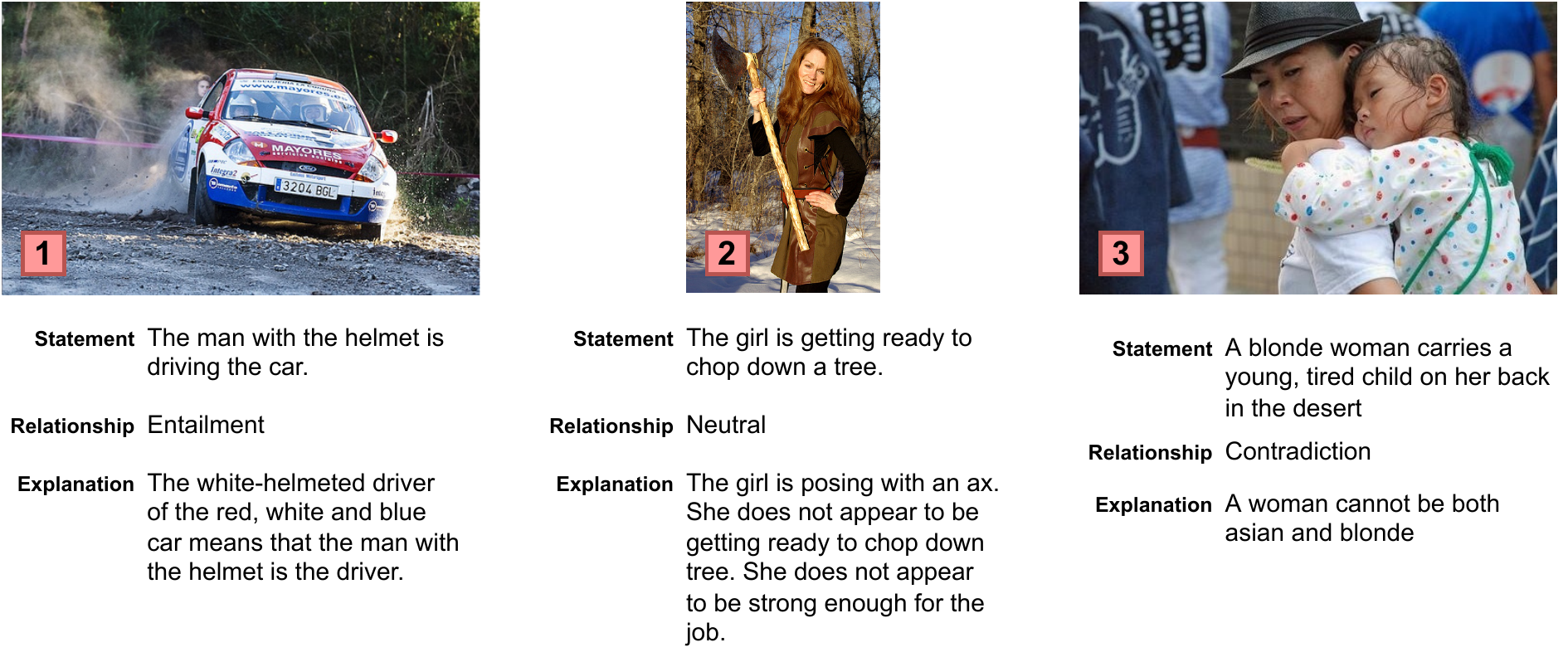}
    \caption{Ethically concerning examples selected from the validation split of the e-SNLI-VE dataset~\cite{kayser_e-vil_2021}.}\label{fig:harmful-bias-esnlive}
\end{figure*}

We present three examples selected from the e-SNLI-VE dataset's validation split that we consider to be ethically problematic due to containing undesired biases in \autoref{fig:harmful-bias-esnlive}. The examples are selected as illustrative instances for ethically concerning data introduced with crowd-sourced vision-language datasets in general.

In example (1), the label ``entailment" for the statement ``The man with the helmet is driving the car" implies that the rally driver is male, although the gender of the driver is not identifiable in the image. In our opinion, this example reflects the societal bias of expecting rally drivers to be male and should be correctly labeled as ``neutral'' with a ground-truth explanation similar to ``The gender of the driver is not clearly identifiable in the image''. In general, similar issues could be avoided by using gender-neutral language where appropriate, which could be implemented in a manual post-processing and moderation step. Example (2) contains a ground-truth explanation that can be considered misogynistic (``she does not appear to be strong enough for the job"). In the third example (3), the explanation ``A woman cannot be both Asian and blonde" represents the wrong belief that Asian women cannot have blonde hair, which is an objectively wrong and arguably racist statement. We further note that on the e-SNLI-VE dataset, people of East Asian ethnicity tend to be labeled as ``Asian''.

Crowd-sourcing datasets inject undesired societal biases and prejudices. Researchers should be aware that models trained on such data reflect these attitudes and can reinforce them when deployed. As a possible remedy, we suggest the implementation of content moderation measures as a post-processing step.

\begin{table*}[t]
  \centering
  \caption{We evaluate OFA-X and NLX-GPT on the first 27 examples from the VQA-X dataset.\label{tab:ofaxvsnlxgpt1}}
  \begin{tabular}{rlp{2.5cm}p{3.4cm}p{3.4cm}p{3.4cm}}
    \textbf{\#} & \textbf{Image}                                                                  & \textbf{Question}                                    & \textbf{OFA-X}                                                                         & \textbf{NLX-GPT}                                                & \textbf{Ground-Truth}                                                                                     \\\\
    \toprule
    0           & \raisebox{-.8\height}{\includegraphics[width=2cm]{figures/ofaxvsnlxgpt/0.png}}  & what is this?                                        & shower because there is a clear glass door                                             & shower because it is a shower stall                             & shower because it has a shower head hanging inside of it                                                  \\\\
    1           & \raisebox{-.8\height}{\includegraphics[width=2cm]{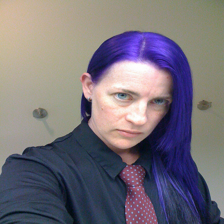}}  & is the woman looking at the camera?                  & yes because her eyes are directed towards the viewer                                   & no because she is not looking at the camera                     & yes because her eyes are open and facing forward                                                          \\\\
    2           & \raisebox{-.8\height}{\includegraphics[width=2cm]{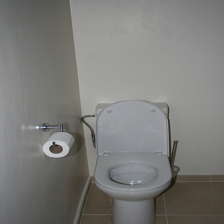}}  & is the toilet clean?                                 & yes because there is no dirt on it                                                     & no because there is dirt all over the seat                      & yes because there is no dirt or marks on the inside or outside                                            \\\\
    3           & \raisebox{-.8\height}{\includegraphics[width=2cm]{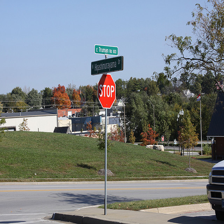}}  & is this a commercial or residential area?            & residential because there are no buildings in this area                                & residential because there are houses and a business             & commercial because there are businesses in the background                                                 \\\\
    4           & \raisebox{-.8\height}{\includegraphics[width=2cm]{figures/ofaxvsnlxgpt/4.png}}  & is the horse jumping?                                & no because its feet are not in the air                                                 & yes because it is galloping                                     & no because all four of its legs are on the ground                                                         \\\\
    5           & \raisebox{-.8\height}{\includegraphics[width=2cm]{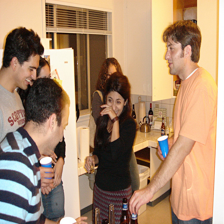}}  & are the people having a party?                       & yes because there are people socializing and enjoying themselves                       & no because there are no people present                          & yes because they are standing in a kitchen with drinks in hand and laughing                               \\\\
    6           & \raisebox{-.8\height}{\includegraphics[width=2cm]{figures/ofaxvsnlxgpt/6.png}}  & what kind of animal is this?                         & giraffe because they have long necks and spots of a giraffe                            & giraffe because it has a long neck and spots                    & giraffe because it has a long neck and is beige with brown spots                                          \\\\
    7           & \raisebox{-.8\height}{\includegraphics[width=2cm]{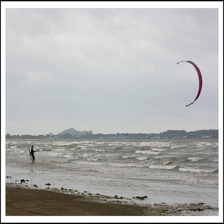}}  & is the water calm?                                   & no because there are many waves in the water                                           & yes because there are waves crashing on the shore               & no because is it windy and there are big waves in the ocean                                               \\\\
    8           & \raisebox{-.8\height}{\includegraphics[width=2cm]{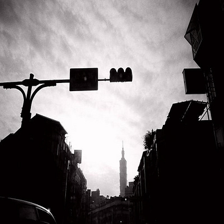}}  & overcast or sunny?                                   & overcast because the sky is covered in clouds                                          & overcast because the sky is dark and the sun is casting shadows & overcast because there are lots of clouds in the sky                                                      
  \end{tabular}
\end{table*}

\begin{table*}[t]
  \centering
  \caption{OFA-X vs NLX-GPT on VQA-X (continued).\label{tab:ofaxvsnlxgpt2}}
  \begin{tabular}{rlp{2.5cm}p{3.4cm}p{3.4cm}p{3.4cm}}
    \textbf{\#} & \textbf{Image}                                                                  & \textbf{Question}                                    & \textbf{OFA-X}                                                                         & \textbf{NLX-GPT}                                                & \textbf{Ground-Truth}                                                                                     \\\\
    \toprule
    9         & \raisebox{-.8\height}{\includegraphics[width=2cm]{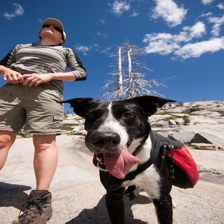}}  & what kind of animal is this?                         & dog because it has a long tongue and is wearing a harness                              & dog because it has four legs and a snout                        & dog because you can tell by its facial features                                                           \\\\
    10        & \raisebox{-.8\height}{\includegraphics[width=2cm]{figures/ofaxvsnlxgpt/10.png}} & what kind of vehicle is shown?                       & bus because it is large and has many seats                                             & bus because it is large and has many seats                      & bus because it is a normal city bus with passengers getting on going to a station                         \\\\
    11        & \raisebox{-.8\height}{\includegraphics[width=2cm]{figures/ofaxvsnlxgpt/11.png}} & is it summer?                                        & no because there is snow                                                               & yes because there is snow on the ground                         & no because there is snow on the ground and people are wearing winter gear                                 \\\\
    12        & \raisebox{-.8\height}{\includegraphics[width=2cm]{figures/ofaxvsnlxgpt/12.png}} & what are the people doing?                           & skiing because they are wearing skis and are on a snowy mountain                       & skiing because they are on skis in the snow                     & skiing because they hold poles and are crossing a snowy mountain                                          \\\\
    13        & \raisebox{-.8\height}{\includegraphics[width=2cm]{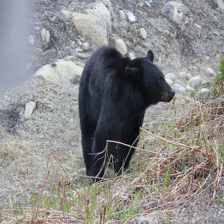}} & does this animal appear to be looking at the camera? & no because the animal is looking to its side                                           & yes because it is looking at something                          & no because the animal is looking off to the side                                                          \\\\
    14        & \raisebox{-.8\height}{\includegraphics[width=2cm]{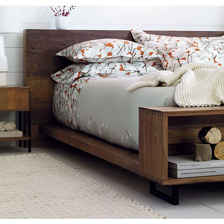}} & is this room tidy?                                   & yes because there are no clutter and all decorative items are neatly arranged          & yes because there are no items on the floor                     & yes because everything is in its place                                                                    \\\\
    15        & \raisebox{-.8\height}{\includegraphics[width=2cm]{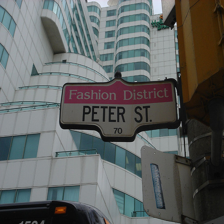}} & what kind of vehicle can be seen?                    & bus because it can hold a lot of people and is long                                    & bus because it is large and has a sign that says "bus stop"     & bus because it is on a street but longer than a car                                                       \\\\
    16        & \raisebox{-.8\height}{\includegraphics[width=2cm]{figures/ofaxvsnlxgpt/16.png}} & what is the gender of the players?                   & male because they are tall and have short hair                                         & male because he is wearing a white shirt and blue shorts        & male because they have masculine builds and are wearing stereotypical male clothing                       \\\\
    17        & \raisebox{-.8\height}{\includegraphics[width=2cm]{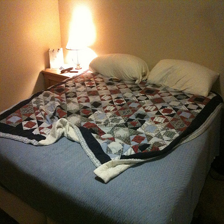}} & is the bed made?                                     & yes because the quilt is folded and the pillows are arranged and the sheets are smooth & yes because the sheets are neatly folded                        & no because the blanket on top has not been pulled all the way flat to cover the entire surface of the bed
  \end{tabular}
\end{table*}

\begin{table*}[t]
  \centering
  \caption{OFA-X vs NLX-GPT on VQA-X (continued).\label{tab:ofaxvsnlxgpt3}}
  \begin{tabular}{rlp{2.5cm}p{3.4cm}p{3.4cm}p{3.4cm}}
    \textbf{\#} & \textbf{Image}                                                                  & \textbf{Question}                                    & \textbf{OFA-X}                                                                         & \textbf{NLX-GPT}                                                & \textbf{Ground-Truth}                                                                                     \\\\
    \toprule
    18        & \raisebox{-.8\height}{\includegraphics[width=2cm]{figures/ofaxvsnlxgpt/18.png}} & is it daytime?                                       & yes because the sky is blue and there are shadows                                      & yes because the sky is blue and there are no clouds             & yes because the large crowd of people are casting shadows from the sun                                    \\\\
    19        & \raisebox{-.8\height}{\includegraphics[width=2cm]{figures/ofaxvsnlxgpt/19.png}} & is it raining?                                       & no because there are no rain drops in the sky                                          & no because the ground is dry and people are using umbrellas     & no because the sky is blue and there isn't a single cloud visible                                         \\\\
    20        & \raisebox{-.8\height}{\includegraphics[width=2cm]{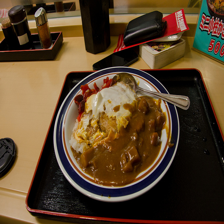}} & is this healthy?                                     & no because there is a lot of fat in that dish                                          & no because there is a lot of fat on the plate                   & no because it consists of particularly greasy and fatty foods                                             \\\\
    21        & \raisebox{-.8\height}{\includegraphics[width=2cm]{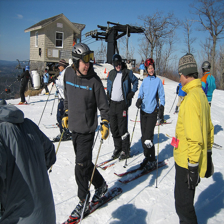}} & is anyone wearing a blue shirt?                      & yes because there is a person in the middle of the picture with a blue coat            & yes because there is a blue stripe on the shirt                 & yes because the shirt they have on is blue                                                                \\\\
    22        & \raisebox{-.8\height}{\includegraphics[width=2cm]{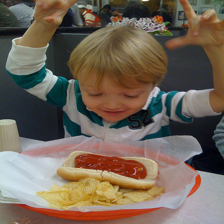}} & does this kid look excited?                          & yes because he is looking at food with excitement                                      & yes because he is smiling and holding a hot dog                 & yes because the boy has his arms in the air and a smile on his face                                       \\\\
    23        & \raisebox{-.8\height}{\includegraphics[width=2cm]{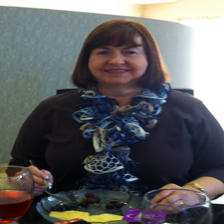}} & what is the woman doing?                             & eating because she has a fork full of food                                             & eating because she is holding a spoon and eating food           & eating because she has a fork touching a plate of food                                                    \\\\
    24        & \raisebox{-.8\height}{\includegraphics[width=2cm]{figures/ofaxvsnlxgpt/24.png}} & is this food healthy?                                & no because it is friday night football                                                 & no because there is a lot of grease on the food truck           & no because potato chips are cooked in oil                                                                 \\\\
    25        & \raisebox{-.8\height}{\includegraphics[width=2cm]{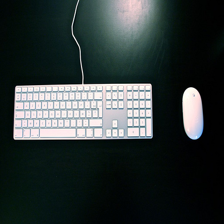}} & is the mouse wireless?                               & yes because there is no cord coming out of it                                          & yes because there is no cord coming out of it                   & yes because there is no cord attached to it                                                               \\\\
    26        & \raisebox{-.8\height}{\includegraphics[width=2cm]{figures/ofaxvsnlxgpt/26.png}} & should the bananas be eaten soon?                    & yes because they are ripe and brown                                                    & yes because they are yellow and ripe                            & yes because spots on them indicate they're getting older                                                  
  \end{tabular}
\end{table*}

\begin{table*}[t]
  \centering
  \caption{Filtered scores for VQA-X, e-SNLI-VE, and VCR. BS stands for BERTScore. For our human results, we report those where participants selected the correct task answer. Note that the code for \cite{majumder_knowledge-grounded_2022} and \cite{palaskar_advances_2022} are not publicly available (see \url{https://github.com/majumderb/rexc/issues/1})}.
  \begin{minipage}[t]{1.0\linewidth}
    \centering
    \label{tab:full-results}
    \begin{tabular}{cccccccccccc}
      \toprule
      \multicolumn{12}{c}{VQA-X}                                                                                                                                                                                 \\
                                            & B1            & B2            & B3            & B4            & R-L           & M             & C              & S             & BS            & Human         & Task Acc.     \\
      \midrule
      PJ-X \cite{park_multimodal_2018}      & 57.4          & 42.4          & 30.9          & 22.7          & 46.0          & 19.7          & 82.7           & 17.1          & 84.6          & 65.4          & 76.4          \\
      FME \cite{wu_faithful_2019}           & 59.1          & 43.4          & 31.7          & 23.1          & 47.1          & 20.4          & 87.0           & 18.4          & 85.2          & 63.2          & 75.5          \\
      \eug \cite{kayser_e-vil_2021}         & 57.3          & 42.7          & 31.4          & 23.2          & 45.7          & 22.1          & 74.1           & 20.1          & 87.0          & 71.5          & 80.5          \\
      \nlxgpt \cite{sammani_nlx-gpt_2022}   & 64.2          & 49.5          & 37.6          & 28.5          & 51.5          & 23.1          & 110.6          & 22.1          & 86.9          & 83.2          & 83.0          \\
      VL-BART \cite{palaskar_advances_2022} & -             & -             & -             & -             & -             & -             & -              & -             & \textbf{91.2} & 75.9          & 86.3          \\
      VL-T5 \cite{palaskar_advances_2022}   & -             & -             & -             & -             & -             & -             & -              & -             & 91.0          & 72.2          & 84.9          \\
      \hline
      OFA-X                                 & \textbf{66.7} & \textbf{51.0} & \textbf{38.5} & \textbf{28.9} & \textbf{52.1} & \textbf{24.5} & \textbf{116.2} & \textbf{24.3} & {87.2}        & \textbf{89.5} & 91.2 \\
      \ofaxmt                               &  64.0        & 49.4           & 37.6          & 28.6          & 51.0          & 23.1          & 110.2         & 22.6          & 86.8          & 87.8       & \textbf{92.6} \\
      \toprule
      \multicolumn{12}{c}{e-SNLI-VE}                                                                                                                                                                                         \\
                                            & B1            & B2            & B3            & B4            & R-L           & M             & C              & S             & BS            & Human         & Task Acc.     \\
      \midrule
      PJ-X                                  & 29.4          & 18.0          & 11.3          & 7.3           & 28.6          & 14.7          & 72.5           & 24.3          & 79.1          & 59.6          & 69.2          \\
      FME                                   & 30.6          & 19.2          & 12.4          & 8.2           & 29.9          & 15.6          & 83.6           & 26.8          & 79.7          & 58.5          & 73.7          \\
      \eug                                  & 30.1          & 19.9          & 13.7          & 9.6           & 27.8          & 19.6 & 85.9           & \textbf{34.5} & 81.7          & 68.9          & 79.5          \\
      \nlxgpt\footnote{NLX-GPT w/ Concepts} & \textbf{37.0} & \textbf{25.3} & \textbf{17.9} & \textbf{12.9} & \textbf{34.2} & 18.8          & \textbf{117.4} & 33.6          & 80.8          & 67.4          & 73.91         \\
      VL-BART                               & -             & -             & -             & -             & -             & -             & -              & -             & \textbf{89.3} & 71.5          & 75,6          \\
      VL-T5                                 & -             & -             & -             & -             & -             & -             & -              & -             & 89.1          & 69.0          & 69.0          \\
      RExC \cite{majumder_knowledge-grounded_2022}                                 & -             & -             & -             & -             & -             &\textbf{22.9}             & -              & -             & 87.7          & 81.8          & 80.8          \\
      \hline
      OFA-X                                 & 34.23         & 22.8          & 15.7          & 11.06         & 32.0          & 18.6          & 115.4          & 33.5          & 80.9          & \textbf{85.7} & \textbf{80.9} \\
      \ofaxmt                               & 32.4         & 21.8           & 15.2          & 10.8          & 31.4          & 17.9          & 108.2         & 32.8          & 80.4          & 78.9       & 78.9 \\
      \toprule
      \multicolumn{12}{c}{VCR}                                                                                                                                                                                               \\
                                            & B1            & B2            & B3            & B4            & R-L           & M             & C              & S             & BS            & Human         & Task Acc.     \\
      \midrule
      PJ-X                                  & 21.8          & 11.0          & 5.9           & 3.4           & 16.4          & 20.5          & 19.0           & 4.5           & 78.4          & 52.7          & 39.0          \\
      FME                                   & 23.0          & 12.5          & 7.2           & 4.4           & 17.3          & 22.7          & 27.7           & \textbf{24.2} & 79.4          & 58.5          & 48.9          \\
      \eug                                  & 20.7          & 11.6          & 6.9           & 4.3           & 11.8          & 22.5          & 32.7           & 12.6          & 79.0          & 65.1          & 69.8          \\
      \nlxgpt                               & \textbf{24.7} & \textbf{15.0} & \textbf{9.6}  & \textbf{6.6}  & 12.2          & \textbf{26.4} & 46.9           & 18.8          & 80.3          & -             & -             \\
      VL-BART                               & -             & -             & -             & -             & -             & -             & -              & -             & \textbf{86.6} & 29.5          & 57.8          \\
      VL-T5                                 & -             & -             & -             & -             & -             & -             & -              & -             & 85.5          & 28.7          & 58.4          \\
      RExC                                  & -             & -             & -             & -             & -             & 20.9             & -              & -             & \textbf{86.6}          & \textbf{80.9}          & \textbf{79.5}          \\
      \hline
      OFA-X                                 & 24.5          & 14.4          & 9.1           & 6.1           & \textbf{25.1} & 12.24         & \textbf{48.5}  & 18.8          & 79.8          & 68.9 & 71.2 \\
      \ofaxmt                               & 22.3         & 13.0           & 8.0          & 5.2          & 24.3          & 11.3          & 44.6         & 17.8         & 79.3          & 77.2       & 62.0 \\
      \bottomrule
    \end{tabular}
  \end{minipage}
\end{table*}

\begin{table*}[]

\centering
\caption{Filtered, unfiltered, and scaled results on selected automatic explanation metrics.}
\label{tab:fitlered-unfiltrd-scaled}
\begin{tabular}{@{}lllllllll@{}}
\toprule
\multicolumn{9}{c}{VQA-X}                                                               \\ 
                         & Method     & B1  & B4  & R-L  & M    & C     & S    & BS   \\ \midrule
\multirow{3}{*}{OFA-X}   & unfiltered & 66.0 & 28.2 & 51.4 & 24.1 & 112.7 & 23.7 & 86.9 \\ 
                         & filtered   & 66.7 & 28.9 & 52.1 & 24.5 & 116.2 & 24.3 & 87.2 \\ 
                         & scaled     & 61.8 & 26.4 & 48.1 & 22.6 & 105.5  & 22.2 & 81.4 \\
                         
                         \midrule
\multirow{3}{*}{\ofaxmt} & unfiltered & 63.3 & 27.9 & 50.5 & 22.7 & 107.4 & 22.1 & 86.6 \\
                         & filtered   & 64.0 & 28.6 & 51.0 & 23.1 & 110.2 & 22.6 & 86.8 \\
                         & scaled     & 59.8 & 26.4 & 47.7 & 21.5 & 101.5 & 20.9 & 81.8 \\

\toprule
\multicolumn{9}{c}{e-SNLI-VE}                                                               \\ 
                         & Method     & B1  & B4  & R-L  & M    & C     & S    & BS   \\ \midrule
\multirow{3}{*}{OFA-X}   & unfiltered & 33.4 & 10.5 & 31.1 & 18.1 & 110.0 & 33.3 & 80.6 \\
                         & filtered   & 34.2 & 11.1 & 32.0 & 18.6 & 115.4 & 33.5 & 80.9 \\
                         & scaled     & 27.1 & 17.8 & 25.2 & 14.7 & 89.0  & 27.0 & 65.2 \\
                         \midrule
                         
\multirow{3}{*}{\ofaxmt} & unfiltered & 31.4 & 10.2 & 30.2 & 17.3 & 101.8 & 32.6 & 80.0 \\ 
                         & filtered   & 32.4 & 10.8 & 31.4 & 17.9 & 108.2 & 32.8 & 80.4 \\ 
                         & scaled     & 24.8 & 8.0  & 23.9 & 13.7 & 80.4  & 25.7 & 63.2 \\
\toprule
\multicolumn{9}{c}{VCR}                                                               \\ 
                         & Method     & B1  & B4  & R-L  & M    & C     & S    & BS   \\ \midrule
\multirow{3}{*}{OFA-X}   & unfiltered & 19.8 &  3.1 & 19.7 &  8.1 & 20.2 & 12.7 & 72.4 \\ 
                         & filtered   & 24.5 &  6.1 & 25.1 & 12.2 & 48.5 & 18.8 & 79.8 \\ 
                         & scaled     & 14.1 &  2.2 & 14.0 &  5.8 & 14.4 &  9.1 & 51.6 \\ 
                         \midrule
\multirow{3}{*}{\ofaxmt} & unfiltered & 20.7 &  3.8 & 19.1 &  7.8 & 19.7 & 12.1 & 72.5 \\ 
                         & filtered   & 22.3 &  5.2 & 24.3 & 11.3 & 44.6 & 17.8 & 79.3 \\
                         & scaled     & 12.8 &  1.9 & 11.9 &  4.8 & 12.2 &  7.5 & 44.9 \\
\bottomrule
\end{tabular}
\end{table*}

\end{document}